\documentclass{article}

\usepackage[final, nonatbib]{neurips_2023}

\usepackage[utf8]{inputenc} % allow utf-8 input
\usepackage[T1]{fontenc}    % use 8-bit T1 fonts
\usepackage{hyperref}       % hyperlinks
\usepackage{url}            % simple URL typesetting
\usepackage{booktabs}       % professional-quality tables
\usepackage{amsfonts}       % blackboard math symbols
\usepackage{nicefrac}       % compact symbols for 1/2, etc.
\usepackage{microtype}      % microtypography
\usepackage{xcolor}         % colors

\usepackage{amsmath}
\usepackage{float}
\usepackage{subcaption}
\usepackage{multicol}
\usepackage[bottom]{footmisc}
\usepackage{diagbox}
\usepackage{multirow, booktabs}
\usepackage{caption}
\usepackage{pifont}% http://ctan.org/pkg/pifont
\newcommand{\xmark}{\ding{55}}%
\usepackage{graphicx}
\newcommand\sbullet[1][.75]{\mathbin{\vcenter{\hbox{\scalebox{#1}{$\bullet$}}}}}
\usepackage{xcolor}
\hypersetup{
    colorlinks=true,
    linkcolor=blue,
    filecolor=blue,      
    urlcolor=blue,
    citecolor=blue,
}

\title{Typhoon: Thai Large Language Models}

\author{%
  \textbf{Kunat Pipatanakul, Phatrasek Jirabovonvisut, Potsawee Manakul} \vspace{1mm} \\
  \textbf{Sittipong Sripaisarnmongkol,  Ruangsak Patomwong} \vspace{1mm} \\ 
  \textbf{Pathomporn Chokchainant, Kasima Tharnpipitchai} \vspace{6.5mm} \\
  SCB 10X \\
}

\begin{document}

\maketitle

\begin{abstract}
    Typhoon is a series of Thai large language models (LLMs) developed specifically for the Thai language. This technical report presents challenges and insights in developing Thai LLMs, including data preparation, pretraining, instruction-tuning, and evaluation. As one of the challenges of low-resource languages is the amount of pretraining data, we apply continual training to transfer existing world knowledge from a strong LLM. To evaluate the Thai knowledge encapsulated in each model from the pretraining stage, we develop ThaiExam, a benchmark based on examinations for high-school students and investment professionals in Thailand. In addition, we fine-tune Typhoon to follow Thai instructions, and we evaluate instruction-tuned models on Thai instruction datasets as well as translation, summarization, and question-answering tasks. Experimental results on a suite of Thai benchmarks show that Typhoon outperforms all open-source Thai language models, and its performance is on par with GPT-3.5 in Thai while having only 7 billion parameters and being 2.62 times more efficient in tokenizing Thai text. 
    
    \vspace{1.5pt}
    \textbf{Model Weights}: \url{https://huggingface.co/scb10x/typhoon-7b} \\
    % \textbf{Webpage}: \url{www.scb10x.com/Typhoon-model-release/} 
    
\end{abstract}

\section{Introduction}
Large Language Models (LLMs) have demonstrated strong zero-shot or few-shot capabilities on a wide range of natural language processing (NLP) tasks \cite{brown2020gpt3, touvron2023llama}. These LLMs are auto-regressive transformers, which are pretrained on a large quantity of self-supervised text data. Most of the major large language models (LLMs) such as GPT-2/3/4~\cite{radford2019language, brown2020gpt3, openai2023gpt4}, Llama-1/2~\cite{touvron2023llama1, touvron2023llama}, Falcon~\cite{penedo2023refinedweb}, Mistral~\cite{jiang2023mistral}, are pretrained primarily on English-centric corpora. In addition, LLMs have been pretrained on specific languages or domains such as Chinese \cite{zeng2023glmb}, finance \cite{wu2023bloomberggpt}, or code \cite{roziere2023code}. 

Given a large number of low-resource languages, multilingual language models have been developed by pretraining a single model on multiple languages. For example, XGLM~\cite{lin-etal-2022-shot} is trained on the CC100 XL corpus, which consists of 30 diverse languages and includes around 11 billion Thai tokens. mT5~\cite{xue-etal-2021-mt5}, a model with an encoder-decoder architecture, is trained on MC4, which covers 101 languages and includes around 11 billion Thai tokens (or 1.14\% of mT5 pretraining data). BLOOM~\cite{workshop2022bloom}, the largest open-source multilingual model with 176 billion parameters, is trained on 46 natural languages (which do not cover Thai) and 13 programming languages. In contrast to high-coverage multilingual LLMs, there exist multilingual LLMs specialized in Southeast Asian (SEA) languages such as Indonesian, Malay, Thai, Vietnamese, and Filipino. For example, SEA-LION~\cite{sea_lion_2023} is a series of multilingual LLMs of different sizes, focusing on SEA languages. However, SEA languages only account for 13\% of the pretraining data of SEA-LION, of which an even smaller part is Thai. Similarly, SeaLLM~\cite{nguyen2023seallms} is another series of LLMs focusing on SEA languages. SeaLLMs are developed by continuing pre-training Llama2 with an extended vocabulary and specialized instruction and alignment tuning. In the pretraining data, Thai text accounts for about 5.91\%. Due to the limited amount of Thai data in pretraining these multilingual models, we argue that they may not have rich Thai knowledge, or understand cultural norms, customs, and stylistic preferences.

The Thai language is spoken by more than 70 million people, but it has received less attention from the NLP community. For instance, the Thai language represents only a small portion of most standard pretraining data. For example, the Thai language constitutes less than 0.5\% of the Common Crawl data~\cite{common_crawl}, making Thai the 26th rank in terms of size in Common Crawl. The Thai language has its own alphabet, which does not overlap with any other high-resource languages. As a result, despite the success of general and multilingual LLMs, we argue that these LLMs may still lack some understanding of Thai culture and knowledge. Therefore, this work focuses on developing LLMs specifically for the Thai language. We hypothesize that a strong LLM can be adapted to the Thai language by pretraining a strong LLM further on a moderately sized Thai corpus.

\begin{figure}[!t]
\includegraphics[width=\linewidth,keepaspectratio]{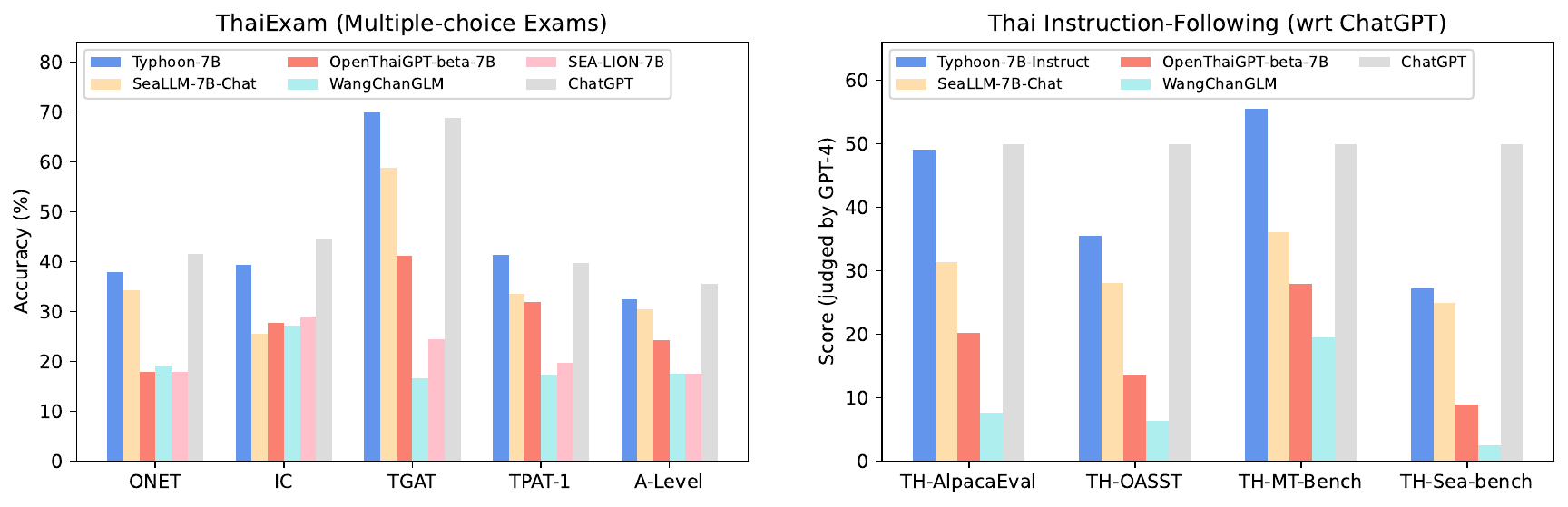}
    \caption{Performance of Typhoon and other open-source Thai large language models on Thai Examinations (left) and Thai instruction-following (right). Details about ThaiExam and Thai instruction-following evaluation are provided in Section~\ref{section:pretrained_eval}, and Section~\ref{section:instruction_eval}, respectively.}
    \label{fig:barplot}
\end{figure}

To this end, this work introduces \textbf{Typhoon}, a 7-billion parameter language model, where its \textit{first} version is adapted from Mistral-7B~\cite{jiang2023mistral}. To assess the extent of Thai knowledge encapsulated within a language model, we develop ThaiExam -- an evaluation benchmark based on Thai examinations. Furthermore, we study fine-tuning Typhoon to follow Thai instructions. We compare instruction-tuned models on both Thai and translated instruction-following datasets, and we also test their zero-shot abilities on machine translation, abstractive summarization, and question-answering tasks. As demonstrated in Figure~\ref{fig:barplot}, the evaluation on various benchmarks shows that Typhoon is the state-of-the-art open-source Thai large language model. Ultimately, this work not only shares our insights from developing Thai large language models but also makes Typhoon publicly available under the Apache-2.0 license to promote further development. 

\section{Related Work}
Training language models on monolingual data (as well as bilingual data where the target language and English data are mixed) has demonstrated gains over multilingual models in both predictive and generative modelling tasks \cite{martin-etal-2020-camembert, nguyen-tuan-nguyen-2020-phobert, sarti2022it5, muller2022cedille, pires2023sabi, sengupta2023jais}. For Thai, WangChangBERTa~\cite{lowphansirikul2021wangchanberta}, an encoder-only model, was the first transformer-based model to be trained on large-scale Thai data. However, as an encoder-only model, WangChangBERTa cannot perform text generation. 

Two open-source Thai generative large language models have also recently been developed. First,  WangChanGLM~\cite{charin_polpanumas_2023_7878101} is based on multilingual XGLM~\cite{lin-etal-2022-shot} with 7.5 billion parameters. WangChanGLM is adapted to Thai by fine-tuning XGLM using LoRA~\cite{hu2021lora} on translated instruction-following datasets of around 380k instruction-response pairs. However, WangChanGLM is not further pretrained on Thai data, so its Thai knowledge is expected to be limited to the pretraining stage of XGLM. Second, OpenThaiGPT (openthaigpt-1.0.0-beta-7b)~\cite{openthaigpt} is based on Llama2~\cite{touvron2023llama} with 7 billion parameters. Its tokenizer extends Llama2's tokenizer to include 24,554 additional Thai tokens to improve generation efficiency. OpenThaiGPT continues pre-training Llama2 on Thai data and performs instruction fine-tuning on the translated instruction datasets. Although the model weights are released to the public, to the best of our knowledge, further details are not publicly available. Additionally, the vanilla Llama2 model may not be able to generate Thai responses, but it can understand Thai to a certain extent (which is shown in our results in Table~\ref{tab:pretrained_model_performance}) due to cross-lingual transferability and the presence of some Thai data during its pretraining. In Table~\ref{tab:thai_llm}, we summarize open-source generative LLMs that are investigated in this work. 

ChatGPT and GPT-4~\cite{openai2023gpt4} are proprietary language model based products that support a variety of languages including Thai. GPT-4 is the state-of-the-art in several English benchmarks~\cite{alpaca_eval, zheng2023judging}. In addition, given the speculated larger amount of pretraining data than any open-source models, ChatGPT and GPT-4 can also be exceptional in low-resource languages.

\begin{table}[!h]
  \centering
  \small
  \begin{tabular}{r|ccccc}
    \toprule
    Model   &MajorLang  &\#Params  &VocabSize &Context &BaseModel    \\
    \midrule
    Typhoon-7B  &English, Thai        &7B  &35,219  &4096   &Mistral   \\
    % Typhoon-34B           &34B &67,514  &4096 &Yi-34B   \\
    OpenThaiGPT-beta-7B &English, Thai        &7B   &56,554 &4096  &Llama2    \\
    WangChanGLM      &Multilingual        &7.5B &256,008 &2048  &XGLM      \\
    % \midrule
    SeaLLM-7B       &Multilingual        &7B   &48,512 &4096  &Llama2  \\
    SEA-LION-7B      &Multilingual        &7B   &256,000 &2048  &\xmark   \\
    % Llama2-13B       &English           &13B  &32,000 &4096   &\xmark &\xmark &2T   &*    \\
    \bottomrule
  \end{tabular}
  \caption{Comparison between Typhoon and open-source language models that support Thai. OpenThaiGPT is based on Llama2 with an extended vocabulary, continual pretraining, and instruction fine-tuning. WangChanGLM is a multi-lingual XGLM model fine-tuned to follow Thai instructions without further pretraining. SEA-LION and SeaLLM are multilingual models specialized in Southeast Asian languages, including Thai. At the time of writing, SEA-LION is only instruction-tuned on Indonesian data, and SeaLLM is only available at 7B size.}
  \label{tab:thai_llm}
\end{table}

\section{Pretraining}
The development of language models tailored to the Thai language presents several unique challenges. Unlike high-resource languages like English, Thai does not have readily available large-scale datasets and resources. This section provides an overview of the challenges and considerations specific to developing Thai LLMs, including data preparation, tokenization, and the choices of base model.

\subsection{Pretraining Data}
In the context of the Thai language, a notable observation emerges: the available Thai internet data constitutes only a small fraction, for instance, less than 0.5\% in the Common Crawl data~\cite{common_crawl}. In addition to the quantity, our initial investigation reveals that there is an issue with the quality where a large portion of crawled Thai texts are search engine optimization (SEO) texts or duplication.

\subsubsection*{Publicly Available Data}
In this work, we consider two primary sources of publicly available Thai data sources as follows. First, MC4~\cite{xue-etal-2021-mt5}, a multilingual cleaned version of Common Crawl's web crawl corpus, consists of texts from 108 languages including Thai. For example, mT5~\cite{xue-etal-2021-mt5} previously used 11 billion Thai tokens in this dataset in its pretraining. Our investigation shows that the quality of MC4 data is higher than Oscar. Second, Oscar~\cite{ortiz-suarez-etal-2020-monolingual}, an open-source multilingual dataset, consists of texts from 151 languages including Thai. The Oscar dataset is larger than MC4, however, we observe that its quality is lower, which necessitates extensive data-cleaning efforts.

\subsubsection*{Methodology}

Given the available resources for pretraining Thai language models, our approach focuses on two key aspects: quantity and quality. In terms of the quantity of data, as the Oscar dataset extracted a single package of the Common Crawl (CC) data and MC4 only made use of a few packages, it is possible to extract more Thai data from the remaining packages of the CC data. In terms of quality, we follow the methodology employed in Falcon RefinedWeb~\cite{penedo2023refinedweb} and apply a deduplication pipeline on Thai data in CC datasets.

Our data preparation can be summarized in four key steps. (1) \textit{Scaling the Data}: we initiate the process by increasing the number of Common Crawl packages where we process a total of 40 packs of the CC data, resulting in a substantially larger dataset comprising approximately 3 TB of Thai text. (2) \textit{Strict Deduplication}: to ensure data quality and minimize redundancy, we implement a deduplication method based on MinHash and LSH algorithms. (3) \textit{Rule-based and Heuristic Filtering}: heuristics are applied to filter data to further enhance data quality. This step involves a manual selection of high-quality websites at the domain level, supplemented by heuristic criteria such as character ratio, line length, and document length. (4) \textit{Incorporating English Data}: to mitigate catastrophic forgetting of the base model's English knowledge \cite{ramasesh2022effect, luo2023empirical}, we add data randomly selected from Falcon RefinedWeb~\cite{penedo2023refinedweb} to our Thai data, resulting in a bilingual dataset with a 50/50 split between Thai and English data. Note that previous work in adapting an LLM to Hungarian shows that mixing English data can mitigate catastrophic forgetting on English, and the adaption is not sensitive to the exact mixture ratio provided that the original language and target language are included \cite{csaki2023efficiently}.

\subsection{Thai Tokenizer}
A byte-level BPE tokenizer, which is adopted by Mistral as well as other LLMs such as GPT or Llama, can encode most words or characters in all languages by falling back to the byte level. Although this tokenizer can operate on a language that it is not initially trained for (e.g., Thai), it has poor efficiency. For instance, previous work shows that the GPT-2 tokenizer with a vocabulary size of 50k requires 3.8 times more tokens to encode Thai texts compared to a tokenizer with a vocabulary size of 5k trained on Thai data~\cite{csaki2023efficiently}. To adapt an existing LLM, the tokenizer of the original model is usually augmented with new tokens where the corresponding embeddings are initialized from scratch~\cite{openthaigpt, csaki2023efficiently}. This results in the trade-off between an improvement in efficiency and the difficulty of model adaption due to new embeddings. Previous work demonstrated that adding beyond 5,000 new Thai tokens yields a diminishing return in efficiency \cite{csaki2023efficiently}.   

In this work, we base our tokenizer on Mistral-7B tokenizer, but we further train an additional Thai subword tokenizer with 5,000 tokens and integrate it with the original tokenizer. This new tokenizer is created by training a SentencePiece model~\cite{kudo-richardson-2018-sentencepiece} on around 8 million samples randomly selected from the Thai subset of MC4 data~\cite{xue-etal-2021-mt5}. We define efficiency as the number of tokens required to represent Thai text with respect to the number of GPT-4 tokens on the same text, i.e.,  
\begin{equation}
\text{Efficiency} = \frac{\#\text{token}_{\text{model}}}{\#\text{token}_{\text{GPT-4}}}
\end{equation}
where $\#\text{token}$ is the number of tokens on the same Thai text. Our experimental results in Table~\ref{tab:efficiency} show that Typhoon's tokenizer is 2.62 times more efficient than GPT-4. 

\begin{table}[!ht]
  \centering
  \small
  \begin{tabular}{c|ccccccc}
    \toprule
    Tokenizer &Typhoon &GPT-4 &Mistral &Character &Word* &OpenThaiGPT &WangChanGLM\\
    \midrule
    Efficiency &262\%  &100\% &93\%  &89\% &326\% &350\% &305\% \\
    \bottomrule
  \end{tabular}
  \caption{Token efficiency with respect to GPT-4. *Based on the \texttt{newmm} tokenizer.}
  
  \label{tab:efficiency}
\end{table}

\subsection{Pretraining Details}
% In this section, we describe our initial investigation and the key findings that will ultimately lead to our final configuration.

\subsubsection*{Initial Experiments}
In the initial experiments, we explore various training techniques and base models on a subset of pretrained data, comprising approximately 500 million tokens. We investigate the following options:

$\sbullet$ \textbf{Base Model}: We compare base models, including Llama2-7B, Llama2-13B~\cite{touvron2023llama}, CodeLlama-34B~\cite{roziere2023code}, and Mistral-7B~\cite{jiang2023mistral} where our criterion is the perplexity. This investigation shows that Mistral-7B achieves the lowest perplexity. Therefore, Mistral-7B is selected as our base model. Note that we perform training on the embeddings and the language modeling head only.

$\sbullet$ \textbf{Training Strategy}:  We compare Low-Rank Adaption (LoRA)~\cite{hu2021lora} and full-weight training on Mistral-7B. Our experiments demonstrate that LoRA yields a lower final loss. Additionally, we tried a two-stage training approach where the embeddings are first trained on full-weight followed by LoRA. However, this setting did not show a significant improvement over using LoRA from the beginning.

$\sbullet$  \textbf{Batch size}: We observed an instability in the training loss when using a batch size of 500k tokens, and we empirically found that increasing the batch size to 2 million tokens can stabilize training.

\subsubsection*{Final Model}

Based on the initial experiments, we continue pretraining the extended embeddings and the language modelling head of Mistral-7B using LoRA. We use the AdamW optimizer~\cite{loshchilov2018decoupled} with a cosine scheduled learning rate initialized at 4e-4. We apply gradient clipping with a threshold of 1.0.

\subsection{Evaluation: Thai Knowledge of Pretrained Models}
\label{section:pretrained_eval}
The NLP community has developed and released challenging, diverse, and holistic benchmarks such as BIG-bench \cite{srivastava2023beyond} and HELM \cite{liang2023holistic}. However, most of these benchmarks focus only on performance in English. Although non-English benchmarks have been proposed, they are still lacking in number \cite{ahuja2023mega, bang2023multitask, zhang2023m3exam}, and only some of these non-English benchmarks such as BHASA \cite{leong2023bhasa} include Thai. Nevertheless, the evaluation datasets in BHASA are still based on multi-lingual datasets or publicly available Thai datasets at a limited scale. To better evaluate the Thai culture and languages of LLMs, we develop \textbf{ThaiExam}, a benchmark comprising Thai multiple-choice examinations as follows:

$\sbullet$ \textbf{ONET}: The Ordinary National Educational Test (ONET) is an examination for students in Thailand. We select the grade-12 ONET exam, which comprises 5 subjects and each question has 5 choices. These subjects are Thai, English, Mathematics, Social Studies, and Science. We extracted these questions from the official 2021 ONET example, amounting to a total of 170 questions and options.

$\sbullet$ \textbf{IC}: The Investment Consultant (IC) examination, a licensing test for investment professionals in Thailand developed by the Stock Exchange of Thailand (SET), features 4 choices per question. We extracted questions for levels 1, 2, and 3 from the official SET website, resulting in a total of 95 questions and options.

$\sbullet$ \textbf{TGAT}: The Thai General Aptitude Test (TGAT), a national high school examination in Thailand, focuses on critical and logical thinking skills, as well as proficiency in the English language. We collected a total of 90 questions and answers. The TGAT consists of four choices per question.

$\sbullet$ \textbf{TPAT-1}: The Thai Professional Aptitude Test 1 (TPAT-1) is a national high school examination in Thailand that assesses students' professional skills requirement in medical schools. This subset contains reasoning and medical ethics. We collected a total of 116 questions and answers. The TPAT-1 consists of 5 choices per question.       

$\sbullet$ \textbf{A-Level}: An academic knowledge assessment examination (Applied Knowledge Level) that covers general foundational subjects taught in schools. The content assessed in this examination aligns with the curriculum guidelines and emphasizes the practical application of knowledge in daily life. We collected a total of 175 questions and answers.

In addition to ThaiExam, we use existing datasets including XNLI (a cross-lingual textual entailment dataset) \cite{conneau-etal-2018-xnli}, XCOPA (a multilingual dataset for causal commonsense reasoning) \cite{ponti-etal-2020-xcopa}, and {M3Exam}~\cite{zhang2023m3exam} (a benchmark developed for evaluating LLMs in a multilingual, multimodal, and multilevel context). The Thai subset of M3Exam is sourced from three levels of ONET: low (grade 6), medium (grade 9), and high (grade 12). 

\begin{table}[!h]
  \centering
  \small
  \tabcolsep=1.1mm
  \begin{tabular}{r|cccccc|ccc}
    \toprule
    \multirow{2}{*}{Model}  &\multicolumn{6}{c}{ThaiExam}   &\multicolumn{3}{c}{Others} \\ 
    
    &ONET &IC &TGAT &TPAT-1 &A-Level &Average &M3Exam &XNLI &XCOPA \\
    \midrule
    Typhoon-7B &\textbf{0.379} &\textbf{0.393} &\textbf{0.700} &\textbf{0.414} &\textbf{0.324} &\textbf{0.442} &\textbf{0.391} &0.421 &\textbf{0.742} \\
    SeaLLM-7B   &0.342 &0.256 &0.589 &0.336 &0.305 &0.366 &0.285 &0.401 &0.586\\
    OpenThaiGPT-beta-7B &0.180 &0.278 &0.411 &0.319 &0.243 &0.286 &0.259 &0.374 &0.570 \\
    WangChanGLM      &0.192 &0.271 &0.167 &0.172 &0.175 &0.195 &0.219 &\textbf{0.431} &0.500\\
    SEA-LION-7B      &0.179 &0.290 &0.244 &0.198 &0.175 &0.217 &0.230 &0.377 &0.462\\
    Llama2-13B       &0.248 &0.241 &0.500 &0.267 &0.273 &0.306 &0.230 &0.366 &0.512 \\
    \midrule
    GPT-3.5-turbo-0613 &0.416 &0.444 &0.689 &0.397 &0.355 &0.460  &0.341$^\dagger$ &0.447 &0.630\\
    GPT-4-0613         &0.531 &0.658 &0.767 &0.491 &0.564 &0.602  &0.560$^\dagger$ &0.623$^\ddagger$ &0.920 \\
    \midrule
    Avg. Human~\cite{onets_stats, tpat1_stats, tcas_stats} &0.318 &- &0.472 &0.406 &- &- &- &- &- \\
    \bottomrule
  \end{tabular}
  \caption{Performance on the Thai reasoning and understanding tasks. $^\dagger$The results are reported in the previous work~\cite{zhang2023m3exam}. $^\ddagger$The experiment was conducted on a subset of XNLI as a full evaluation would cost more than \$1000.}
\label{tab:pretrained_model_performance}
\end{table}

In Table~\ref{tab:pretrained_model_performance}, we compare our pre-trained Typhoon against open-source and proprietary LLMs that support the Thai language. The results show that Typhoon-7B is the best-performing model among all open-source models across the evaluation datasets. Typhoon-7B also outperforms an average human tester on ONET, TGAT and TPAT-1 examinations. When compared to proprietary (and potentially much larger) models, Typhoon despite having only 7 billion parameters outperforms GPT-3.5 on 4 out of 8 evaluation datasets. 

\section{Instruction-Tuning}
Although pretrained language models are highly capable of predicting the next words, even large language models of more than 100 billion parameters may still lack the ability to follow a user's intent well. As a result, instruction-tuning has been applied to align language models with a user's intent \cite{ouyang2022training, alpaca, touvron2023llama}. Instruction-tuning or fine-tuning methods could be categorized into supervised fine-tuning (SFT) and alignment via reinforcement learning using human feedback (RLHF) \cite{stiennon2020learning, ouyang2022training} or AI feedback (RLAIF) \cite{lee2023rlaif}, and Direct Preference Optimization (DPO)~\cite{rafailov2023direct}. However, these techniques require significant annotation resources, which do not yet exist in the Thai language.  

% results & evaluation (if we have)
\subsection{Supervised Fine-Tuning (SFT)}
To address the lack of instruction-tuning data, we consider three directions as follows: (1) translating English instruction-tuning datasets into Thai, (2) converting Thai NLP datasets using pre-defined prompt templates similar to the FLAN collection \cite{wei2022finetuned, chung2022scaling}, (3) generating Thai instruction-tuning data using the Self-Instruct technique \cite{wang-etal-2023-self-instruct}.

As a first step, we create SFT data by translating English instruction-following datasets into Thai. The translation-based approach was also previously used in fine-tuning WangChanGLM where datasets such as Alpaca~\cite{alpaca} and Dolly~\cite{DatabricksBlog2023DollyV2} were translated using Google Translate API. In this work, we investigate translation further by using Typhoon fine-tuned for translation and GPT-4; for example, we translate UltraChat~\cite{ding2023enhancing_ultrachat} using GPT-4. Additionally, we follow self-instruct~\cite{wang-etal-2023-self-instruct} in generating a Thai-instruction dataset. Using translated and self-instruct data, we obtain Typhoon-7B-Instruct by performing supervised fine-tuning.

\subsection{Evaluation: Instruction-Following Tasks}
\label{section:instruction_eval}
In Section~\ref{section:pretrained_eval}, we evaluate Thai knowledge using multi-choice question answering and classification benchmarks. However, the previous evaluation may not reflect how well the models understand human intents, and the evaluation of instruction-following models typically requires human interactions~\cite{alpaca_eval}. Therefore, we perform instruction-following evaluation using the following datasets:

$\sbullet$ \textbf{Thai AlpacaEval}: We create an instruction-following dataset following AlpacaEval~ \cite{alpaca_eval}. As native Thai speakers, we \textit{manually} write prompts in Thai based on the examples randomly selected in AlpacaEval. For instance, given an example from AlpacaEval, we write a prompt in Thai as well as using a Thai context instead of a literal translation. This dataset consists of 105 Thai prompts. 

$\sbullet$ \textbf{Thai OASST}: We make use of a Thai subset of OpenAssistant (OASST1), 166 human-written instruction-response pairs, which was previously used in evaluating WangChanGLM. 

$\sbullet$ \textbf{Translated MT-Bench}: We translate MT-Bench~\cite{zheng2023judging} (using Google API) into Thai and we filter out poorly translated examples such as coding instructions, resulting in a total of 68 examples. The translated MT-Bench is used to evaluate the multi-turn ability of the models.

$\sbullet$ \textbf{Sea-bench (Thai subset)}~\cite{nguyen2023seallms}: We make use of a Thai subset of Sea-bench inspired by MT-Bench for SEA languages released with SeaLLM. It utilizes translations of English test sets, real user questions from local sources, math and reasoning questions, and linguist-written instructions. Sea-bench encompasses various categories, including task-solving, math-reasoning, general instructions, NaturalQA for colloquial language, and safety instructions. This dataset consists of 100 Thai prompts.

We use an LLM as a judge to perform pairwise comparison \cite{alpaca_eval, zheng2023judging, liusie2023llm} by measuring the fraction of times the LLM judge (e.g., GPT-4) prefers the responses from the model being assessed over the responses from a reference model. The LLM judge compares each pair twice in a symmetric manner, and if the outputs of the LLM judge are not the same, it is treated as a draw, i.e., the model being evaluated gets a score of 0.5 (\textit{tie}). Otherwise, the model either gets a score of 0.0 (\textit{loss}) or 1.0 (\textit{win}).

In this experiment, we use GPT-4 (gpt-4-1106) as the LLM judge, and the reference model is GPT-3.5 (gpt-3.5-turbo-0613) as the reference model such that it would obtain a score of 0.50. For single-turn evaluation sets (e.g., Thai-Alpaca, Thai-OASST, and Sea-bench), we follow the LLM-judge prompt adopted by AlpacaEval.\footnote{\url{https://github.com/tatsu-lab/alpaca_eval/blob/main/src/alpaca_eval/evaluators_configs/chatgpt/basic_prompt.txt}} For multi-turn evaluation sets (e.g., translated MT-Bench), we follow the LLM-judge prompt adopted in FastChat.\footnote{\url{https://github.com/lm-sys/FastChat/blob/main/fastchat/llm_judge/data/judge_prompts.jsonl}}

\begin{table}[!h]
  \centering
  \small
  \begin{tabular}{r|cccc}
    \toprule
    Model  &AlpacaEval &OASST  &MT-Bench  &Sea-bench  \\
    \midrule
    % Typhoon-7B           &Pretrained     &19.52  &17.77   &14.62   \\
    % SEA-LION-7B       &Pretrained      &4.76 &5.72 &22.06     \\
    % \midrule
    Typhoon-7B-Instruct       &\textbf{49.05}     &\textbf{35.54} &\textbf{55.52}  &\textbf{27.25} \\ % instruct-v6
    SeaLLM-7B-Chat             &31.43     &28.01   &36.03   &25.00  \\
    % SEA-LION-7B      &Pretrained         & & & \\
    OpenThaiGPT-beta-7B      &20.24     &13.55   &27.94   &9.00  \\
    % WangChan-LION-7B      &14.05	&14.16	&20.96	&12.00    \\
    WangChanGLM           &7.62      &6.33    &19.49  &2.50    \\
    % SeaLLM-13B-Chat  &Instruct           &*     &*   &*   &49.75  \\ % model weight not released yet
    \bottomrule
  \end{tabular}
  \caption{Instruction-following evaluation (Score in \%) based on Thai AlpacaEval, Thai OASST, translated MT-Bench and Sea-bench. Each model is compared against GPT-3.5 using GPT-4 as the judge, and the metric is win-rate defined as the number of times that the judge prefers the responses from the model over the responses from the reference model using symmetric pairwise comparison.}
  \label{tab:instruction_tuning_performance}
\end{table}

In Table~\ref{tab:instruction_tuning_performance}, we compare Typhoon-7B-Instruct against other instruction-following models. At the time of conducting the experiments, SEA-LION is only instruction-tuned in Indonesian. GPT-3.5 is used as a reference model and GPT-4 is used as a judge, and hence are not shown in Table~\ref{tab:instruction_tuning_performance}. Our experimental results show that Typhoon-7B-Instruct is comparable to GPT-3.5 on Thai AlpacaEval and translated MT-Bench but it is worse on Thai OASST and Sea-bench. When compared to open-source Thai LLMs, Typhoon-7B-Instruct achieves state-of-the-art performance on all instruction-following datasets.

\subsection{Evaluation: Translation, Summarization, Question-Answering}
\label{section:nlp_tasks}
In addition to instruction-tuning evaluation with an LLM judge, we evaluate the LLMs on standard Natural Language Processing (NLP) tasks, including machine translation, abstractive summarization, and question-answering. Here, we investigate the zero-shot abilities of Thai LLMs. First, we evaluate the English-to-Thai machine translation performance on FLORES-200 \cite{costa2022no}. Second, we evaluate the abstractive summarization performance on XLSum (Thai article $\rightarrow$ Thai summary) \cite{hasan-etal-2021-xl} and CrossSum (English article $\rightarrow$ Thai summary) \cite{bhattacharjee-etal-2023-crosssum}. Third, we evaluate the question-answering performance on XQuAD (i.e., reading comprehension task) \cite{artetxe-etal-2020-cross}. Our experimental results in Table~\ref{tab:nlp_performance} show that Typhoon-7B-Instruct is the best-performing system on most of the NLP tasks investigated.

\begin{table}[!h]
  \centering
  \small
  \tabcolsep=1.7mm
  \begin{tabular}{r|cccc}
    \toprule
    \multirow{2}{*}{Model}   &Translation   &\multicolumn{2}{c}{Summarization}  &Question-Answering \\  
    &FLORES-200 &XLSum(Th$\rightarrow$Th) &CrossSum(En$\rightarrow$Th) &XQuAD (Th) \\
    \midrule
    Typhoon-7B-Instruct       &\textbf{31.14}/\textbf{46.62} &\textbf{21.60}/\textbf{4.24}/\textbf{14.51}   &17.32/\textbf{3.83}/11.39  &34.46/\textbf{54.03} \\ % instruct-v6
    SeaLLM-7B-Chat &14.36/37.13 &10.21/1.72/7.96 &15.66/2.33/10.58   &20.89/47.95  \\
    % SEA-LION-7B    \\
    OpenThaiGPT-beta-7B      &27.68/44.41 &17.98/1.93/12.69   &17.72/3.06/11.28  &\textbf{34.58}/40.82 \\
    % WangChan-LION-7B      &20.41/36.40 &13.40/3.51/10.89 &20.59/\textbf{4.18}/\textbf{16.17} &7.37/24.95 \\
    WangChanGLM           &10.63/31.78 &19.10/2.10/13.77   &\textbf{22.91}/3.76/\textbf{15.61} &17.81/19.94 \\
    \bottomrule
  \end{tabular}
  \caption{Zero-shot performance on NLP tasks. Machine translation is evaluated on FLORES-200 using BLEU \cite{papineni-etal-2002-bleu} and chrF \cite{popovic-2015-chrf}. Summarization is evaluated on XLSum (Thai-to-Thai) and CrossSum (English-to-Thai) using ROUGE-1/2/L \cite{lin-2004-rouge} with the \texttt{newmm} tokenizer. Question-Answering is evaluated on XQuAD using word-overlap F1 metric in 0-shot and 1-shot settings, respectively.}
  \label{tab:nlp_performance}
\end{table}

\section{Risk and Limitations} % Safety considerations
Similar to other language models, Typhoon may (i) hallucinate, e.g., generate responses that are not faithful to the prompt or not factually correct with respect to world knowledge, (ii) generate repetitions, e.g., repeated words, phrases or sentences, (iii) produce harmful or inappropriate responses.  

\section{Conclusion and Future Work}
Our work on Typhoon, a Thai large language model with 7 billion parameters, demonstrates that we can adapt an existing English-centric LLM to Thai using only a subset of Thai data that we currently have. Typhoon is the state-of-the-art open-source model on Thai benchmarks as well as achieving performance on par with GPT-3.5 in Thai while being 2.62 times more efficient in tokenization. Future work will extend pretraining to utilize a larger amount of Thai data that is available, and use larger base models such as 34B, 70B, or mixture-of-experts to exploit the emergent ability. Also, future work will investigate instruction tuning further for an improved alignment.

\section*{Acknowledgements}
We would like to thank Mukaya Panich, Tanwa Arpornthip, Unnawut Leepaisalsuwanna, and Panuwat Chayabunjonglerd for their advice on this project. We would like to thank the SCBX R\&D team for their support with the evaluation. We would also like to thank Sarana Nutanong and the NLP team at VISTEC for their feedback on the experiments and this technical report.

\bibliographystyle{plain}  % apalike
\bibliography{anthology, references}

\begin{thebibliography}{10}

\bibitem{ahuja2023mega}
Kabir Ahuja, Rishav Hada, Millicent Ochieng, Prachi Jain, Harshita Diddee, Samuel Maina, Tanuja Ganu, Sameer Segal, Maxamed Axmed, Kalika Bali, et~al.
\newblock Mega: Multilingual evaluation of generative ai.
\newblock {\em arXiv preprint arXiv:2303.12528}, 2023.

\bibitem{artetxe-etal-2020-cross}
Mikel Artetxe, Sebastian Ruder, and Dani Yogatama.
\newblock On the cross-lingual transferability of monolingual representations.
\newblock In {\em Proceedings of the 58th Annual Meeting of the Association for Computational Linguistics}, pages 4623--4637, Online, July 2020. Association for Computational Linguistics.

\bibitem{bang2023multitask}
Yejin Bang, Samuel Cahyawijaya, Nayeon Lee, Wenliang Dai, Dan Su, Bryan Wilie, Holy Lovenia, Ziwei Ji, Tiezheng Yu, Willy Chung, Quyet~V. Do, Yan Xu, and Pascale Fung.
\newblock A multitask, multilingual, multimodal evaluation of chatgpt on reasoning, hallucination, and interactivity.
\newblock {\em arXiv preprint arXiv:2302.04023}, 2023.

\bibitem{srivastava2023beyond}
BIG bench authors.
\newblock Beyond the imitation game: Quantifying and extrapolating the capabilities of language models.
\newblock {\em Transactions on Machine Learning Research}, 2023.

\bibitem{bhattacharjee-etal-2023-crosssum}
Abhik Bhattacharjee, Tahmid Hasan, Wasi~Uddin Ahmad, Yuan-Fang Li, Yong-Bin Kang, and Rifat Shahriyar.
\newblock {C}ross{S}um: Beyond {E}nglish-centric cross-lingual summarization for 1,500+ language pairs.
\newblock In Anna Rogers, Jordan Boyd-Graber, and Naoaki Okazaki, editors, {\em Proceedings of the 61st Annual Meeting of the Association for Computational Linguistics (Volume 1: Long Papers)}, pages 2541--2564, Toronto, Canada, July 2023. Association for Computational Linguistics.

\bibitem{brown2020gpt3}
Tom Brown, Benjamin Mann, Nick Ryder, Melanie Subbiah, Jared~D Kaplan, Prafulla Dhariwal, Arvind Neelakantan, Pranav Shyam, Girish Sastry, Amanda Askell, Sandhini Agarwal, Ariel Herbert-Voss, Gretchen Krueger, Tom Henighan, Rewon Child, Aditya Ramesh, Daniel Ziegler, Jeffrey Wu, Clemens Winter, Chris Hesse, Mark Chen, Eric Sigler, Mateusz Litwin, Scott Gray, Benjamin Chess, Jack Clark, Christopher Berner, Sam McCandlish, Alec Radford, Ilya Sutskever, and Dario Amodei.
\newblock Language models are few-shot learners.
\newblock In H.~Larochelle, M.~Ranzato, R.~Hadsell, M.F. Balcan, and H.~Lin, editors, {\em Advances in Neural Information Processing Systems}, volume~33, pages 1877--1901. Curran Associates, Inc., 2020.

\bibitem{chung2022scaling}
Hyung~Won Chung, Le~Hou, Shayne Longpre, Barret Zoph, Yi~Tay, William Fedus, Yunxuan Li, Xuezhi Wang, Mostafa Dehghani, Siddhartha Brahma, et~al.
\newblock Scaling instruction-finetuned language models.
\newblock {\em arXiv preprint arXiv:2210.11416}, 2022.

\bibitem{conneau-etal-2018-xnli}
Alexis Conneau, Ruty Rinott, Guillaume Lample, Adina Williams, Samuel Bowman, Holger Schwenk, and Veselin Stoyanov.
\newblock {XNLI}: Evaluating cross-lingual sentence representations.
\newblock In {\em Proceedings of the 2018 Conference on Empirical Methods in Natural Language Processing}, pages 2475--2485, Brussels, Belgium, October-November 2018. Association for Computational Linguistics.

\bibitem{DatabricksBlog2023DollyV2}
Mike Conover, Matt Hayes, Ankit Mathur, Jianwei Xie, Jun Wan, Sam Shah, Ali Ghodsi, Patrick Wendell, Matei Zaharia, and Reynold Xin.
\newblock Free dolly: Introducing the world's first truly open instruction-tuned llm, 2023.

\bibitem{costa2022no}
Marta~R Costa-jussa, James Cross, Onur {\c{C}}elebi, Maha Elbayad, Kenneth Heafield, Kevin Heffernan, Elahe Kalbassi, Janice Lam, Daniel Licht, Jean Maillard, et~al.
\newblock No language left behind: Scaling human-centered machine translation.
\newblock {\em arXiv preprint arXiv:2207.04672}, 2022.

\bibitem{csaki2023efficiently}
Zoltan Csaki, Pian Pawakapan, Urmish Thakker, and Qiantong Xu.
\newblock Efficiently adapting pretrained language models to new languages.
\newblock {\em arXiv preprint arXiv:2311.05741}, 2023.

\bibitem{ding2023enhancing_ultrachat}
Ning Ding, Yulin Chen, Bokai Xu, Yujia Qin, Shengding Hu, Zhiyuan Liu, Maosong Sun, and Bowen Zhou.
\newblock Enhancing chat language models by scaling high-quality instructional conversations.
\newblock In Houda Bouamor, Juan Pino, and Kalika Bali, editors, {\em Proceedings of the 2023 Conference on Empirical Methods in Natural Language Processing}, pages 3029--3051, Singapore, December 2023. Association for Computational Linguistics.

\bibitem{common_crawl}
Common~Crawl Foundation.
\newblock Statistics of common crawl monthly archives by commoncrawl.
\newblock https://commoncrawl.github.io/cc-crawl-statistics/plots/languages, 2023.

\bibitem{hasan-etal-2021-xl}
Tahmid Hasan, Abhik Bhattacharjee, Md.~Saiful Islam, Kazi Mubasshir, Yuan-Fang Li, Yong-Bin Kang, M.~Sohel Rahman, and Rifat Shahriyar.
\newblock {XL}-sum: Large-scale multilingual abstractive summarization for 44 languages.
\newblock In {\em Findings of the Association for Computational Linguistics: ACL-IJCNLP 2021}, pages 4693--4703, Online, August 2021. Association for Computational Linguistics.

\bibitem{hu2021lora}
Edward~J Hu, yelong shen, Phillip Wallis, Zeyuan Allen-Zhu, Yuanzhi Li, Shean Wang, Lu~Wang, and Weizhu Chen.
\newblock Lo{RA}: Low-rank adaptation of large language models.
\newblock In {\em International Conference on Learning Representations}, 2022.

\bibitem{jiang2023mistral}
Albert~Q Jiang, Alexandre Sablayrolles, Arthur Mensch, Chris Bamford, Devendra~Singh Chaplot, Diego de~las Casas, Florian Bressand, Gianna Lengyel, Guillaume Lample, Lucile Saulnier, et~al.
\newblock Mistral 7b.
\newblock {\em arXiv preprint arXiv:2310.06825}, 2023.

\bibitem{kudo-richardson-2018-sentencepiece}
Taku Kudo and John Richardson.
\newblock {S}entence{P}iece: A simple and language independent subword tokenizer and detokenizer for neural text processing.
\newblock In {\em Proceedings of the 2018 Conference on Empirical Methods in Natural Language Processing: System Demonstrations}, pages 66--71, Brussels, Belgium, November 2018. Association for Computational Linguistics.

\bibitem{lee2023rlaif}
Harrison Lee, Samrat Phatale, Hassan Mansoor, Kellie Lu, Thomas Mesnard, Colton Bishop, Victor Carbune, and Abhinav Rastogi.
\newblock Rlaif: Scaling reinforcement learning from human feedback with ai feedback.
\newblock {\em arXiv preprint arXiv:2309.00267}, 2023.

\bibitem{leong2023bhasa}
Wei~Qi Leong, Jian~Gang Ngui, Yosephine Susanto, Hamsawardhini Rengarajan, Kengatharaiyer Sarveswaran, and William~Chandra Tjhi.
\newblock Bhasa: A holistic southeast asian linguistic and cultural evaluation suite for large language models.
\newblock {\em arXiv preprint arXiv:2309.06085}, 2023.

\bibitem{alpaca_eval}
Xuechen Li, Tianyi Zhang, Yann Dubois, Rohan Taori, Ishaan Gulrajani, Carlos Guestrin, Percy Liang, and Tatsunori~B. Hashimoto.
\newblock Alpacaeval: An automatic evaluator of instruction-following models.
\newblock \url{https://github.com/tatsu-lab/alpaca_eval}, 2023.

\bibitem{liang2023holistic}
Percy Liang, Rishi Bommasani, Tony Lee, Dimitris Tsipras, Dilara Soylu, Michihiro Yasunaga, Yian Zhang, Deepak Narayanan, Yuhuai Wu, Ananya Kumar, Benjamin Newman, Binhang Yuan, Bobby Yan, Ce~Zhang, Christian Cosgrove, Christopher~D. Manning, Christopher Ré, Diana Acosta-Navas, Drew~A. Hudson, Eric Zelikman, Esin Durmus, Faisal Ladhak, Frieda Rong, Hongyu Ren, Huaxiu Yao, Jue Wang, Keshav Santhanam, Laurel Orr, Lucia Zheng, Mert Yuksekgonul, Mirac Suzgun, Nathan Kim, Neel Guha, Niladri Chatterji, Omar Khattab, Peter Henderson, Qian Huang, Ryan Chi, Sang~Michael Xie, Shibani Santurkar, Surya Ganguli, Tatsunori Hashimoto, Thomas Icard, Tianyi Zhang, Vishrav Chaudhary, William Wang, Xuechen Li, Yifan Mai, Yuhui Zhang, and Yuta Koreeda.
\newblock Holistic evaluation of language models, 2023.

\bibitem{lin-2004-rouge}
Chin-Yew Lin.
\newblock {ROUGE}: A package for automatic evaluation of summaries.
\newblock In {\em Text Summarization Branches Out}, pages 74--81, Barcelona, Spain, July 2004. Association for Computational Linguistics.

\bibitem{lin-etal-2022-shot}
Xi~Victoria Lin, Todor Mihaylov, Mikel Artetxe, Tianlu Wang, Shuohui Chen, Daniel Simig, Myle Ott, Naman Goyal, Shruti Bhosale, Jingfei Du, Ramakanth Pasunuru, Sam Shleifer, Punit~Singh Koura, Vishrav Chaudhary, Brian O{'}Horo, Jeff Wang, Luke Zettlemoyer, Zornitsa Kozareva, Mona Diab, Veselin Stoyanov, and Xian Li.
\newblock Few-shot learning with multilingual generative language models.
\newblock In {\em Proceedings of the 2022 Conference on Empirical Methods in Natural Language Processing}, pages 9019--9052, Abu Dhabi, United Arab Emirates, December 2022. Association for Computational Linguistics.

\bibitem{liusie2023llm}
Adian Liusie, Potsawee Manakul, and Mark J.~F. Gales.
\newblock {LLM} {C}omparative {A}ssessment: {Z}ero-shot {NLG} {E}valuation through {P}airwise {C}omparisons using {L}arge {L}anguage {M}odels, 2023.

\bibitem{loshchilov2018decoupled}
Ilya Loshchilov and Frank Hutter.
\newblock Decoupled weight decay regularization.
\newblock In {\em International Conference on Learning Representations}, 2019.

\bibitem{lowphansirikul2021wangchanberta}
Lalita Lowphansirikul, Charin Polpanumas, Nawat Jantrakulchai, and Sarana Nutanong.
\newblock Wangchanberta: Pretraining transformer-based thai language models.
\newblock {\em arXiv preprint arXiv:2101.09635}, 2021.

\bibitem{luo2023empirical}
Yun Luo, Zhen Yang, Fandong Meng, Yafu Li, Jie Zhou, and Yue Zhang.
\newblock An empirical study of catastrophic forgetting in large language models during continual fine-tuning.
\newblock {\em arXiv preprint arXiv:2308.08747}, 2023.

\bibitem{martin-etal-2020-camembert}
Louis Martin, Benjamin Muller, Pedro~Javier Ortiz~Su{\'a}rez, Yoann Dupont, Laurent Romary, {\'E}ric de~la Clergerie, Djam{\'e} Seddah, and Beno{\^\i}t Sagot.
\newblock {C}amem{BERT}: a tasty {F}rench language model.
\newblock In {\em Proceedings of the 58th Annual Meeting of the Association for Computational Linguistics}, pages 7203--7219, Online, July 2020. Association for Computational Linguistics.

\bibitem{muller2022cedille}
Martin M{\"u}ller and Florian Laurent.
\newblock Cedille: A large autoregressive french language model.
\newblock {\em arXiv preprint arXiv:2202.03371}, 2022.

\bibitem{nguyen-tuan-nguyen-2020-phobert}
Dat~Quoc Nguyen and Anh Tuan~Nguyen.
\newblock {P}ho{BERT}: Pre-trained language models for {V}ietnamese.
\newblock In {\em Findings of the Association for Computational Linguistics: EMNLP 2020}, pages 1037--1042, Online, November 2020. Association for Computational Linguistics.

\bibitem{nguyen2023seallms}
Xuan-Phi Nguyen, Wenxuan Zhang, Xin Li, Mahani Aljunied, Qingyu Tan, Liying Cheng, Guanzheng Chen, Yue Deng, Sen Yang, Chaoqun Liu, et~al.
\newblock Seallms--large language models for southeast asia.
\newblock {\em arXiv preprint arXiv:2312.00738}, 2023.

\bibitem{onets_stats}
The National~Institute of~Educational Testing~Service.
\newblock Basic statistical values of o-net test results.
\newblock \url{https://www.niets.or.th/th/content/view/11821}, 2021.

\bibitem{tpat1_stats}
Consortium of~Thai Medical~Schools.
\newblock Scores report of the tpat1 exam for thai medical schools admission.
\newblock \url{https://www9.si.mahidol.ac.th/cotmes_stat.html}, 2023.

\bibitem{tcas_stats}
Council of~University Presidents~of Thailand.
\newblock Basic statistical report tgat/tpat examination.
\newblock \url{https://www.mytcas.com/stat/}, 2023.

\bibitem{openai2023gpt4}
OpenAI.
\newblock {GPT}-4 technical report, 2023.

\bibitem{openthaigpt}
Open{T}hai{GPT}.
\newblock Released openthaigpt 7b 1.0.0-beta.
\newblock \url{https://openthaigpt.aieat.or.th/}, 2023.

\bibitem{ortiz-suarez-etal-2020-monolingual}
Pedro~Javier Ortiz~Su{\'a}rez, Laurent Romary, and Beno{\^\i}t Sagot.
\newblock A monolingual approach to contextualized word embeddings for mid-resource languages.
\newblock In {\em Proceedings of the 58th Annual Meeting of the Association for Computational Linguistics}, pages 1703--1714, Online, July 2020. Association for Computational Linguistics.

\bibitem{ouyang2022training}
Long Ouyang, Jeffrey Wu, Xu~Jiang, Diogo Almeida, Carroll Wainwright, Pamela Mishkin, Chong Zhang, Sandhini Agarwal, Katarina Slama, Alex Ray, et~al.
\newblock Training language models to follow instructions with human feedback.
\newblock {\em Advances in Neural Information Processing Systems}, 35:27730--27744, 2022.

\bibitem{papineni-etal-2002-bleu}
Kishore Papineni, Salim Roukos, Todd Ward, and Wei-Jing Zhu.
\newblock {B}leu: a method for automatic evaluation of machine translation.
\newblock In {\em Proceedings of the 40th Annual Meeting of the Association for Computational Linguistics}, pages 311--318, Philadelphia, Pennsylvania, USA, July 2002. Association for Computational Linguistics.

\bibitem{penedo2023refinedweb}
Guilherme Penedo, Quentin Malartic, Daniel Hesslow, Ruxandra Cojocaru, Alessandro Cappelli, Hamza Alobeidli, Baptiste Pannier, Ebtesam Almazrouei, and Julien Launay.
\newblock The refinedweb dataset for falcon llm: Outperforming curated corpora with web data, and web data only, 2023.

\bibitem{pires2023sabi}
Ramon Pires, Hugo Abonizio, Thales Rog{\'e}rio, and Rodrigo Nogueira.
\newblock Sabi$\backslash$'a: Portuguese large language models.
\newblock {\em arXiv preprint arXiv:2304.07880}, 2023.

\bibitem{charin_polpanumas_2023_7878101}
Charin Polpanumas, Wannaphong Phatthiyaphaibun, Patomporn Payoungkhamdee, Peerat Limkonchotiwat, Lalita Lowphansirikul, Can Udomcharoenchaikit, Titipat Achakulwisut, Ekapol Chuangsuwanich, and Sarana Nutanong.
\newblock {WangChanGLM — The Multilingual Instruction-Following Model}, April 2023.

\bibitem{ponti-etal-2020-xcopa}
Edoardo~Maria Ponti, Goran Glava{\v{s}}, Olga Majewska, Qianchu Liu, Ivan Vuli{\'c}, and Anna Korhonen.
\newblock {XCOPA}: A multilingual dataset for causal commonsense reasoning.
\newblock In {\em Proceedings of the 2020 Conference on Empirical Methods in Natural Language Processing (EMNLP)}, pages 2362--2376, Online, November 2020. Association for Computational Linguistics.

\bibitem{popovic-2015-chrf}
Maja Popovi{\'c}.
\newblock chr{F}: character n-gram {F}-score for automatic {MT} evaluation.
\newblock In {\em Proceedings of the Tenth Workshop on Statistical Machine Translation}, pages 392--395, Lisbon, Portugal, September 2015. Association for Computational Linguistics.

\bibitem{radford2019language}
Alec Radford, Jeffrey Wu, Rewon Child, David Luan, Dario Amodei, Ilya Sutskever, et~al.
\newblock Language models are unsupervised multitask learners.
\newblock {\em OpenAI blog}, 1(8):9, 2019.

\bibitem{rafailov2023direct}
Rafael Rafailov, Archit Sharma, Eric Mitchell, Stefano Ermon, Christopher~D Manning, and Chelsea Finn.
\newblock Direct preference optimization: Your language model is secretly a reward model.
\newblock {\em arXiv preprint arXiv:2305.18290}, 2023.

\bibitem{ramasesh2022effect}
Vinay~Venkatesh Ramasesh, Aitor Lewkowycz, and Ethan Dyer.
\newblock Effect of scale on catastrophic forgetting in neural networks.
\newblock In {\em International Conference on Learning Representations}, 2022.

\bibitem{roziere2023code}
Baptiste Roziere, Jonas Gehring, Fabian Gloeckle, Sten Sootla, Itai Gat, Xiaoqing~Ellen Tan, Yossi Adi, Jingyu Liu, Tal Remez, J{\'e}r{\'e}my Rapin, et~al.
\newblock Code llama: Open foundation models for code.
\newblock {\em arXiv preprint arXiv:2308.12950}, 2023.

\bibitem{sarti2022it5}
Gabriele Sarti and Malvina Nissim.
\newblock It5: Large-scale text-to-text pretraining for italian language understanding and generation.
\newblock {\em arXiv preprint arXiv:2203.03759}, 2022.

\bibitem{sengupta2023jais}
Neha Sengupta, Sunil~Kumar Sahu, Bokang Jia, Satheesh Katipomu, Haonan Li, Fajri Koto, Osama~Mohammed Afzal, Samta Kamboj, Onkar Pandit, Rahul Pal, et~al.
\newblock Jais and jais-chat: Arabic-centric foundation and instruction-tuned open generative large language models.
\newblock {\em arXiv preprint arXiv:2308.16149}, 2023.

\bibitem{sea_lion_2023}
AI~Singapore.
\newblock Sea-lion (southeast asian languages in one network): A family of large language models for southeast asia.
\newblock \url{https://github.com/aisingapore/sealion}, 2023.

\bibitem{stiennon2020learning}
Nisan Stiennon, Long Ouyang, Jeffrey Wu, Daniel Ziegler, Ryan Lowe, Chelsea Voss, Alec Radford, Dario Amodei, and Paul~F Christiano.
\newblock Learning to summarize with human feedback.
\newblock {\em Advances in Neural Information Processing Systems}, 33:3008--3021, 2020.

\bibitem{alpaca}
Rohan Taori, Ishaan Gulrajani, Tianyi Zhang, Yann Dubois, Xuechen Li, Carlos Guestrin, Percy Liang, and Tatsunori~B. Hashimoto.
\newblock Stanford alpaca: An instruction-following llama model.
\newblock \url{https://github.com/tatsu-lab/stanford_alpaca}, 2023.

\bibitem{touvron2023llama1}
Hugo Touvron, Thibaut Lavril, Gautier Izacard, Xavier Martinet, Marie-Anne Lachaux, Timoth{\'e}e Lacroix, Baptiste Rozi{\`e}re, Naman Goyal, Eric Hambro, Faisal Azhar, et~al.
\newblock Llama: Open and efficient foundation language models.
\newblock {\em arXiv preprint arXiv:2302.13971}, 2023.

\bibitem{touvron2023llama}
Hugo Touvron, Louis Martin, Kevin Stone, Peter Albert, Amjad Almahairi, Yasmine Babaei, Nikolay Bashlykov, Soumya Batra, Prajjwal Bhargava, Shruti Bhosale, et~al.
\newblock Llama 2: Open foundation and fine-tuned chat models.
\newblock {\em arXiv preprint arXiv:2307.09288}, 2023.

\bibitem{wang-etal-2023-self-instruct}
Yizhong Wang, Yeganeh Kordi, Swaroop Mishra, Alisa Liu, Noah~A. Smith, Daniel Khashabi, and Hannaneh Hajishirzi.
\newblock Self-instruct: Aligning language models with self-generated instructions.
\newblock In Anna Rogers, Jordan Boyd-Graber, and Naoaki Okazaki, editors, {\em Proceedings of the 61st Annual Meeting of the Association for Computational Linguistics (Volume 1: Long Papers)}, pages 13484--13508, Toronto, Canada, July 2023. Association for Computational Linguistics.

\bibitem{wei2022finetuned}
Jason Wei, Maarten Bosma, Vincent Zhao, Kelvin Guu, Adams~Wei Yu, Brian Lester, Nan Du, Andrew~M. Dai, and Quoc~V Le.
\newblock Finetuned language models are zero-shot learners.
\newblock In {\em International Conference on Learning Representations}, 2022.

\bibitem{workshop2022bloom}
BigScience Workshop, Teven~Le Scao, Angela Fan, Christopher Akiki, Ellie Pavlick, Suzana Ili{\'c}, Daniel Hesslow, Roman Castagn{\'e}, Alexandra~Sasha Luccioni, Fran{\c{c}}ois Yvon, et~al.
\newblock Bloom: A 176b-parameter open-access multilingual language model.
\newblock {\em arXiv preprint arXiv:2211.05100}, 2022.

\bibitem{wu2023bloomberggpt}
Shijie Wu, Ozan Irsoy, Steven Lu, Vadim Dabravolski, Mark Dredze, Sebastian Gehrmann, Prabhanjan Kambadur, David Rosenberg, and Gideon Mann.
\newblock Bloomberggpt: A large language model for finance.
\newblock {\em arXiv preprint arXiv:2303.17564}, 2023.

\bibitem{xue-etal-2021-mt5}
Linting Xue, Noah Constant, Adam Roberts, Mihir Kale, Rami Al-Rfou, Aditya Siddhant, Aditya Barua, and Colin Raffel.
\newblock m{T}5: A massively multilingual pre-trained text-to-text transformer.
\newblock In {\em Proceedings of the 2021 Conference of the North American Chapter of the Association for Computational Linguistics: Human Language Technologies}, pages 483--498, Online, June 2021. Association for Computational Linguistics.

\bibitem{zeng2023glmb}
Aohan Zeng, Xiao Liu, Zhengxiao Du, Zihan Wang, Hanyu Lai, Ming Ding, Zhuoyi Yang, Yifan Xu, Wendi Zheng, Xiao Xia, Weng~Lam Tam, Zixuan Ma, Yufei Xue, Jidong Zhai, Wenguang Chen, Zhiyuan Liu, Peng Zhang, Yuxiao Dong, and Jie Tang.
\newblock {GLM}-130b: An open bilingual pre-trained model.
\newblock In {\em The Eleventh International Conference on Learning Representations}, 2023.

\bibitem{zhang2023m3exam}
Wenxuan Zhang, Sharifah~Mahani Aljunied, Chang Gao, Yew~Ken Chia, and Lidong Bing.
\newblock M3exam: A multilingual, multimodal, multilevel benchmark for examining large language models.
\newblock {\em arXiv preprint arXiv:2306.05179}, 2023.

\bibitem{zheng2023judging}
Lianmin Zheng, Wei-Lin Chiang, Ying Sheng, Siyuan Zhuang, Zhanghao Wu, Yonghao Zhuang, Zi~Lin, Zhuohan Li, Dacheng Li, Eric Xing, et~al.
\newblock Judging llm-as-a-judge with mt-bench and chatbot arena.
\newblock {\em arXiv preprint arXiv:2306.05685}, 2023.

\end{thebibliography}

%%%%%%%%%%%%%%%%%%%%%%%%%%%%%%%%%%%%%%%%%%%%%%%%%%%%%%%%%%%%

\end{document}